\documentclass{article}

\usepackage[preprint, nonatbib]{neurips_2023}

\usepackage[utf8]{inputenc} 
\usepackage[T1]{fontenc}    
\usepackage{hyperref}       
\usepackage{url}            
\usepackage{booktabs}       
\usepackage{amsfonts}       
\usepackage{nicefrac}       
\usepackage{microtype}      
\usepackage{times}
\usepackage{epsfig}
\usepackage{graphicx}
\usepackage{mathtools}
\usepackage{amssymb,amsmath,amsthm}

\usepackage{algorithm}
\usepackage{multirow}
\usepackage{subfigure}
\usepackage{tikz} 
\usetikzlibrary{arrows}
\usepackage{bbm}
\usepackage{svg}

\title{AdaGossip: Adaptive Consensus Step-size for Decentralized Deep Learning with Communication Compression}

\author{%
  Sai Aparna Aketi \hspace{2mm} Abolfazl Hashemi \hspace{2mm} Kaushik Roy\\
  Department of Electrical and Computer Engineering\\
  Purdue University\\
  West Lafayette, IN 47906 \\
  \texttt{\{saketi, abolfazl, kaushik\}@purdue.edu} \\
}

\usepackage{xr}
\makeatletter

\newcommand*{\addFileDependency}[1]{
\typeout{(#1)}
\@addtofilelist{#1}
\IfFileExists{#1}{}{\typeout{No file #1.}}
}\makeatother

\newcommand*{\myexternaldocument}[1]{%
\externaldocument{#1}%
\addFileDependency{#1.tex}%
\addFileDependency{#1.aux}%
}

\myexternaldocument{main_appendix}

\begin{document}

\maketitle

\begin{abstract}
Decentralized learning is crucial in supporting on-device learning over large distributed datasets, eliminating the need for a central server. However, the communication overhead remains a major bottleneck for the practical realization of such decentralized setups. To tackle this issue, several algorithms for decentralized training with compressed communication have been proposed in the literature. Most of these algorithms introduce an additional hyper-parameter referred to as consensus step-size which is tuned based on the compression ratio at the beginning of the training. 
In this work, we propose AdaGossip, a novel technique that adaptively adjusts the consensus step-size based on the compressed model differences between neighboring agents. 
We demonstrate the effectiveness of the proposed method through an exhaustive set of experiments on various Computer Vision datasets (CIFAR-10, CIFAR-100, Fashion MNIST, Imagenette, and ImageNet), model architectures, and network topologies. Our experiments show that the proposed method achieves superior performance ($0-2\%$ improvement in test accuracy) compared to the current state-of-the-art method for decentralized learning with communication compression. 

\end{abstract}

\section{Introduction}
The remarkable success of deep learning models is largely attributed to the availability of large amounts of data and computational power \cite{vit, gpt4}. Traditionally, these models were trained on a single system or a cluster of nodes by centralizing data from various distributed sources. 
This necessitates the transfer of a tremendous amount of information from edge devices to the central server, incurring significant power consumption, and an increase in users' privacy concerns. 
To tackle this challenge, a new interest in developing distributed learning has emerged \cite{konevcny2016federated}.  

Decentralized learning is a branch of distributed optimization that focuses on learning from data distributed across multiple agents without a central server. 
It offers many advantages over the traditional centralized approach in core aspects such as data privacy, fault tolerance, and scalability \cite{nedic2020distributed}.
Studies have shown that decentralized learning algorithms \cite{d-psgd} can perform comparable to centralized algorithms on benchmark vision datasets.
However, in the realm of Foundational Models and Large Language Models with billions of model parameters, communication overhead becomes one of the major bottlenecks for the practical realization of decentralized learning setups. 
To achieve communication efficiency in decentralized learning, several techniques employing error-feedback based compression have been proposed in the literature  \cite{deepsqueeze, quant_sgp, choco-sgd, moniqua, sparsepush, lp, powergossip, beer}.  

Deep-Squeeze \cite{deepsqueeze} introduces error-compensated communication compression to a gossip-based decentralized algorithm namely Decentralized Parallel Stochastic Gradient Descent (DSGD) \cite{d-psgd}. In particular, Deep-squeeze compresses error-compensated model parameters before communicating them to the neighboring agents. 
In \cite{choco1, choco-sgd}, authors proposed the CHOCO-SGD algorithm which allows for arbitrarily high communication compression leveraging an error feedback mechanism while allowing the nodes to communicate the compressed model differences. Communicating model differences instead of model parameters, at the cost of additional memory requirement, helps CHOCO-SGD achieve better trade-offs between the compression ratio and performance drop. Quantized decentralized stochastic learning algorithm over directed graphs is introduced in \cite{quant_sgp}. This work combines CHOCO-SGD with Stochastic Gradient Push (SGP) \cite{sgp} to enable communication compression over directed graphs. Sparse-Push \cite{sparsepush} extends Deep-Squeeze to time-varying and directed graphs and analyses the impact of heterogeneous data distributions on communication compression. PowerGossip \cite{powergossip}, inspired by PowerSGD \cite{powersgd}, compresses the model differences between neighboring agents using low-rank linear compressors. However, the experimental results are shown on a small scale of 8 workers and the labeled images are reshuffled between 8 workers every epoch which isn't practical. BEER \cite{beer} adopts communication compression with gradient tracking showing a faster convergence rate of $\mathcal{O}(1/T)$. However, it has been shown that gradient tracking doesn't scale well for deep neural networks \cite{qgm}. 
Most of the above-mentioned algorithms introduce an additional hyper-parameter referred to as consensus step-size. The tuning of this hyper-parameter becomes crucial in achieving competitive performance with a full-communication baseline.

In this paper, we present AdaGossip, a novel decentralized method that dynamically adjusts the consensus step-size based on the compressed model differences between neighboring agents. 
Here, the average of the compressed model differences between neighboring agents is referred to as the gossip-error. 
Inspired by AdaGrad \cite{adagrad}, AdaGossip computes individual adaptive consensus step-size for different parameters from the estimates of second moments of the gossip-error. 
The main intuition behind the AdaGossip is that the higher gossip-error for a given parameter can indicate a higher impact of the compression requiring a lower consensus step-size. Therefore, utilizing the dynamics of gossip-error to adjust the consensus step-size gives the flexibility for each model parameter to be averaged at a different rate depending on the impact of the (biased) compressor on them.
We validate the effectiveness of the proposed algorithm through an exhaustive set of experiments on various datasets, model architectures, compressors, and graph topologies. 
In particular, we show that integrating AdaGossip with state-of-the-art communication compression methods such as CHOCO-SGD improves the performance of decentralized learning. 

\subsection{Contributions}
In summary, we make the following contributions.
\begin{itemize}
    \item We Propose AdaGossip, a novel decentralized learning method that dynamically adjusts the consensus step-size based on the compressed model differences between neighboring agents. 
    \item We present AdaG-SGD that extends the proposed AdaGossip to decentralized machine learning setups.
    \item The exhaustive set of experiments on various datasets, model architectures, compressors, and graph topologies establish that the proposed AdaG-SGD improves the performance of decentralized learning with communication compression. 
 .
\end{itemize}

\section{Background}

In this section, we provide the background on the decentralized setup with communication compression.

The main goal of decentralized machine learning is to learn a global model using the knowledge extracted from the locally stored data samples across $n$ agents while maintaining privacy constraints. In particular, we solve the optimization problem of minimizing the global loss function $f(x)$ distributed across $n$ agents as given in \eqref{eq:1}. 
Note that $F_i$ is a local loss function (for example, cross-entropy loss) defined in terms of the data ($d_i$) sampled from the local dataset $D_i$ at agent $i$ and $x_i$ is the model parameters vector of agent $i$.
\begin{equation}
\label{eq:1}
\begin{split}
    \min \limits_{x \in \mathbb{R}^d} f(x) &= \frac{1}{n}\sum_{i=1}^n f_i(x), \\
    \text{where} \hspace{2mm} f_i(x) &= \mathbb{E}_{d_i \sim D_i}[F_i(x; d_i)], \hspace{2mm} \text{for all } i.
\end{split}
\end{equation}
The optimization problem is typically solved by combining stochastic gradient descent \cite{sgd} with global consensus-based gossip averaging \cite{gossip}. 
The communication topology is modeled as a graph $G = ([n], E)$ with edges $\{i,j\} \in E$ if and only if agents $i$ and $j$ are connected by a communication link exchanging the messages directly. 
We represent $\mathcal{N}(i)$ as the neighbors of agent $i$ including itself. It is assumed that the graph $G$ is strongly connected with self-loops 
i.e., there is a path from every agent to every other agent. 
The adjacency matrix of the graph $G$ is referred to as a mixing matrix $W$ where $w_{ij}$ is the weight associated with the edge $\{i,j\}$. Note that, weight $0$ indicates the absence of a direct edge between the agents, and the elements of the Identity matrix are represented by $I_{ij}$.
Similar to the majority of previous works in decentralized learning, the mixing matrix is assumed to be doubly stochastic. 
Further, the initial models and all the hyperparameters are synchronized at the beginning of the training. The communication among the agents is assumed to be synchronous. 

\begin{algorithm}[ht]
\textbf{Input:} Each agent $i \in [1,n]$ initializes model weights $x^0_{i}$, learning rate $\eta$, consensus step-size $\gamma$, mixing matrix $W=[w_{ij}]_{i,j \in [1,n]}$.\\

Each agent simultaneously implements the T\text{\scriptsize RAIN}( ) procedure\\
\textbf{procedure} T\text{\scriptsize RAIN}( ) \\
\hspace*{4mm}\textbf{for} t=$0,1,\hdots,T-1$ \textbf{do}\\
\hspace*{8mm}$d^t_{i} \sim D_i$\hfill\textcolor{gray!80}{// sample data from training dataset.}\\
\hspace*{8mm}$g^{t}_{i}=\nabla_x F_i(d^t_{i}; x_i^{t}) $ \hfill \textcolor{gray!80}{// compute the local gradients.}\\
\hspace*{8mm}$x^{t+\frac{1}{2}}_{i}=x^t_{i} - \eta g^t_{i}$\hfill\textcolor{gray!80}{// update the model.}\\
\hspace*{8mm}S\text{\scriptsize END}R\text{\scriptsize ECEIVE}($x^{t+\frac{1}{2}}_{i}$)\hfill\textcolor{gray!80}{// share model parameters with neighbors.}\\
\hspace*{8mm}$x^{(t+1)}_{i}=x^{t+\frac{1}{2}}_{i}+\gamma\sum_{j\in \mathcal{N}_i}w_{ij}(x^{t+\frac{1}{2}}_{j} - x^{t+\frac{1}{2}}_{i})$\hfill \textcolor{gray!80}{// gossip averaging step.}\\
\hspace*{4mm}\textbf{end}\\
\textbf{return}
\caption{{Decentralized SGD (\textit{DSGD})}}
\label{alg:dl}
\end{algorithm}

Traditional decentralized algorithms referred to as Decentralized Stochastic Gradient Descent (DSGD) is \cite{d-psgd} given by Algorithm.~\ref{alg:dl}. DSGD assumes the data across the agents to be Independent and Identically Distributed (IID) and communicates the model parameters at every iteration. Note that the consensus step-size $\gamma$ for DSGD is set to one and line-7 of Algorithm.~\ref{alg:dl} can be simplified as $x^{(t+1)}_{i}=\sum_{j\in \mathcal{N}_i}w_{ij} x^{t+\frac{1}{2}}_{j}$. The matrix notation of DSGD is given by Equation.~\eqref{eq:dsgd}.
\begin{equation}
\label{eq:dsgd}
\begin{split}
     X^{(t+1)} &= X^{t} - \eta G^t + \gamma X^t (W-I)
\end{split}
\end{equation}
for iterates $X^t := [x_1^t, ..., x_n^t] \in \mathbb{R}^{d \times n}$ and $G^t := [\nabla F_1(x_1; d_1), ..., \nabla F_n(x_n; d_n)] \in \mathbb{R}^{d \times n}$. $I$ is the identity matrix of size $n \times n$.

To achieve communication efficiency, algorithms such as DeepSqueeze \cite{deepsqueeze}, CHOCO-SGD \cite{choco-sgd} etc., incorporate error-compensated communication compression into DSGD. DeepSqueeze applies the compression operator ($C_\omega$) to the error-compensated model parameters ($V^t$) as shown in Equation.~\eqref{eq:ds}. The error compensation is computed by the variable $\Delta^t$ which is the difference between the true message and its compressed version.
\begin{equation}
\label{eq:ds}
\begin{split}
     \text{DeepSqueeze}: X^{(t+1)} &= X^{t} - \eta G^t + \gamma C_\omega [V^t] \hspace{1mm} (W-I)\\
     V^t & = X^{t} - \eta G^t + \Delta^t \hspace{1mm}; \hspace{2mm} \Delta^t = V^{t-1}- C_\omega [V^{t-1}]
\end{split}
\end{equation}
In contrast, CHOCO-SGD \cite{choco-sgd} compresses and communicates the difference between the updated model parameters ($X^t - \eta G^t)$  and its "publicly available" copy ($\hat{X}^t$). Each agent $i \in [n]$ maintains and updates its own model parameters $x_i$ along with the variable $\hat{x}_j$ for all neighbors, including itself. $\hat{x}_i$ is referred to as the publicly available copy of the private $x_i$ as it is available to all the neighbors of $i$ and is updated based on the received (compressed) information. In general $x_i \neq \hat{x}_i$, due to communication compression.
\begin{equation}
\label{eq:choco}
\begin{split}
     \text{CHOCO-SGD}: X^{(t+1)} &= X^{t} - \eta G^t + \gamma \hat{X}^{t+1} \hspace{1mm} (W-I)\\
     \hat{X}^{t+1} & = \hat{X}^{t} + \Delta^{t} \hspace{1mm}; \hspace{2mm} \Delta^t = C_\omega [X^{t} - \eta G^t - \hat{X}^{t}]
\end{split}
\end{equation}

Both algorithms carefully tune the consensus step-size ($\gamma$) based on the compression ratio. Note that even though PowerGossip \cite{powergossip} and BEER \cite{beer} claim better performance than CHOCO-SGD for simpler tasks, their scalability to larger graph structures and deep neural networks respectively is not guaranteed and needs further study. Hence, we compare our method with CHOCO-SGD which is the current state-of-the-art for decentralized deep learning with communication compression. 

\section{AdaGossip}
In this section, we outline a novel gossip algorithm called AdaGossip, and its adaptation to decentralized learning referred to as AdaG-SGD. AdaGossip introduces adaptive consensus step-size to a compressed gossip algorithm aiming to improve performance under constrained communication. Algorithm.~\ref{alg:adagossip} summarizes the proposed AdaGossip utilizing CHOCO-Gossip \cite{choco1} as the base communication compression algorithm. 

\begin{algorithm}[ht]
\textbf{Input:} Each agent $i \in [1,n]$  has initial data $x^0_{i} \in \mathbb{R}^d$, initializes $\hat{x}_j^0 = 0 \hspace{2mm}\forall j \in \mathcal{N}_i$, $u_i^0 = 0$. Mixing matrix $W=[w_{ij}]_{i,j \in [1,n]}$, Compression operator $C_\omega$. Hyper-parameters: $\gamma \in (0,1]$, $\beta \in [0,1)$. $(e_i^t)^2$ indicates the elementwise square $e_i^t \odot e_i^t$.\\

Each agent simultaneously implements the following\\
\hspace{4mm}\textbf{for} t=$0,1,\hdots,T-1$ \textbf{do}\\
\hspace*{6mm}$\delta_i^t = C_\omega[x_i^t - \hat{x}_i^t]$\\
\hspace*{6mm}S\text{\scriptsize END}R\text{\scriptsize ECEIVE}($\delta^t$)\\
\hspace*{6mm}\textbf{for} each neighbor $j \in \{\mathcal{N}_i\}$ \textbf{do}\\
\hspace*{10mm}$\hat{x}_j^{t+1}=\hat{x}_j^t+\delta^t_j$\\
\hspace*{6mm}\textbf{end}\\
\hspace*{6mm}$e^{t}_{i}=\sum_{j\in \mathcal{N}_i}w_{ij}(\hat{x}^{t+1}_{j} - \hat{x}^{t+1}_{i})$\hfill\textcolor{gray!80}{// compute gossip-error.}\\
\hspace*{6mm}$u^{t}_{i}=\beta u_i^{t-1} + (1-\beta).(e_i^t)^2$\hfill\textcolor{gray!80}{// second raw moment estimate of gossip-error.}\\
\hspace*{6mm}$x^{(t+1)}_{i}=x^{t}_{i}+\frac{\gamma}{\sqrt{u_i^t}+\epsilon} e_i^t$\hfill\textcolor{gray!80}{// gossip averaging step with adaptive consensus step-size.}\\
\hspace{4mm}\textbf{end}\\
\caption{AdaGossip}
\label{alg:adagossip}
\end{algorithm}

Traditionally, gossip-type Distributed Average Consensus algorithms with communication compression such as CHOCO-Gossip \cite{choco1}  introduce an additional hyper-parameter referred to as consensus step-size ($\gamma$). This hyper-parameter determines the rate at which the received information (model parameters of neighboring models) is averaged.
Now, in the case of communication compression, the averaging of compressed neighboring models happens at a lower rate ($\gamma <1$) as the received information is erroneous. 
Therefore tuning consensus step-size based on the compression ratio becomes crucial in achieving competitive performance with a full communication baseline. To address this, we propose AdaGossip which adaptively determines the consensus step-size for each parameter based on its variation in the compressed models of the neighboring agents. 

AdaGossip, described in Algorithm.~\ref{alg:adagossip}, adjusts the individual adaptive consensus step-size ($\gamma_i^t$) for different parameters from the estimates of the second raw moments of the gossip-error. Here, gossip-error ($e_i$) for a given agent $i$ is defined as the average difference between the compressed model of the neighboring agents ($\hat{x}_j$) and the compressed local model ($\hat{x}_i$). 
The second raw moment of the gossip-error given by $u_i$ is the exponential moving average of the squared (element-wise) gossip-error where $\beta \in [0,1)$ controls the exponential decay rate.
A high gossip-error for a given parameter indicates a high error in the received neighbors' parameter due to compression and hence needs to be averaged at a lower rate. AdaGossip accommodates this knowledge by scaling the consensus step-size proportional to the second raw moment of the gossip-error. In particular, the consensus step-size at a given iteration $t$ on agent $i$ is computed as $\gamma_i^t = \gamma/(\sqrt{u_i^t}+\epsilon)$ where the hyper-parameter $\gamma \in (0,1]$.

\begin{algorithm}[ht]
\textbf{Input:} Each agent $i \in [1,n]$ initializes: model weights $x^0_{i} \in \mathbb{R}^d$, $\hat{x}_j^0 = 0 \hspace{2mm}\forall j \in \mathcal{N}_i$, $u_i^0 = 0$. Mixing matrix $W=[w_{ij}]_{i,j \in [1,n]}$, Compression operator $C_\omega$. Hyper-parameters: learning rate $\eta$, $\gamma \in (0,1]$, $\beta \in [0,1)$. $(e_i^t)^2$ indicates the elementwise square $e_i^t \odot e_i^t$.\\

Each agent simultaneously implements the T\text{\scriptsize RAIN}( ) procedure\\
\textbf{procedure} T\text{\scriptsize RAIN}( ) \\
\hspace*{4mm}\textbf{for} t=$0,1,\hdots,T-1$ \textbf{do}\\
\hspace*{8mm}Sample $d^t_{i} \sim D_i$ and compute gradient $g^{t}_{i}=\nabla_x F_i(d^t_{i}; x_i^{t}) $ \\
\hspace*{8mm} $x^{t+\frac{1}{2}}_{i}=x^t_{i} - \eta g^t_{i}$ \hfill\textcolor{gray!80}{// update the model.}\\
\hspace*{8mm}$\delta_i^t = C_\omega[x_i^{t+\frac{1}{2}} - \hat{x}_i^t]$ \hfill\textcolor{gray!80}{// compress error-compensated model update.}\\
\hspace*{8mm}S\text{\scriptsize END}R\text{\scriptsize ECEIVE}($\delta^t$)\hfill\textcolor{gray!80}{// share compressed variable with neighbors.}\\
\hspace*{8mm}\textbf{for} each neighbor $j \in \{\mathcal{N}_i\}$ \textbf{do}\\
\hspace*{12mm}$\hat{x}_j^{t+1}=\hat{x}_j^t+\delta^t_j$\hfill\textcolor{gray!80}{// recover model parameters of neighbors.}\\
\hspace*{8mm}\textbf{end}\\
\hspace*{8mm}$e^{t}_{i}=\sum_{j\in \mathcal{N}_i}w_{ij}(\hat{x}^{t+1}_{j} - \hat{x}^{t+1}_{i})$\hfill\textcolor{gray!80}{// compute gossip-error.}\\
\hspace*{8mm}$u^{t}_{i}=\beta u_i^{t-1} + (1-\beta).(e_i^t)^2$\hfill\textcolor{gray!80}{// second raw moment estimate of gossip-error.}\\
\hspace*{8mm}$x^{(t+1)}_{i}=x^{t}_{i}+\frac{\gamma}{\sqrt{u_i^t}+\epsilon} e_i^t$\hfill\textcolor{gray!80}{// gossip averaging step with adaptive consensus step-size.}\\
\hspace*{4mm}\textbf{end}\\
\textbf{return}
\caption{AdaG-SGD}
\label{alg:AdagSGD}
\end{algorithm}

We further extend the proposed AdaGossip Algorithm to decentralized machine learning referred to as \emph{AdaG-SGD}. Algorithm.\ref{alg:AdagSGD} outlines the proposed AdaG-SGD which is a decentralized learning algorithm with communication compression and adaptive consensus step-size. 


\section{Experiments}

In this section, we demonstrate the performance of the proposed \textit{AdaG-SGD} technique compared with the current state-of-the-art CHOCO-SGD \cite{choco-sgd}. \footnote{The PyTorch code is available at \url{https://github.com/aparna-aketi/AdaG_SGD}}.

\subsection{Experimental Setup}
We analyze the efficiency of the proposed methods through experiments on a diverse set of datasets, model architectures, graph topologies, and graph sizes. We present the analysis on -- 
(a) \textbf{Datasets:} CIFAR-10, CIFAR-100, Fashion MNIST, Imagenette and ImageNet.
(b) \textbf{Model architectures:} ResNet, LeNet-5 and, MobileNet-V2. All the models use Evonorm \cite{evonorm} as the activation-normalization layer as it is shown to be better suited for decentralized learning on heterogeneous data.
(c) \textbf{Graph topologies:} Ring graph with 2 peers per agent, Dyck graph with 3 peers per agent, and Torus graph with 4 peers per agent.
(d) \textbf{Number of agents:} 8-40 agents.

The graph topologies used for the decentralized setup are undirected and have a uniform mixing matrix. The undirected ring topology for any graph size has 3 peers per agent including itself and each edge weights $\frac{1}{3}$. The undirected Dyck topology with 32 agents has 4 peers per agent including itself and each edge weights $\frac{1}{4}$. The undirected torus topology with 32 agents has 5 peers per agent including itself and each edge weights $\frac{1}{5}$. 
The data is distributed across the agents in an IID manner similar to \cite{deepsqueeze, choco-sgd, powergossip}. 
The partitioned data is fixed, non-overlapping, and never shuffled across agents during the training. 

The initial learning rate is either set to 0.1 (CIFAR-10, CIFAR-100, ImageNet) or 0.01 (Fashion MNIST, Imagenette) and is decayed by a factor of 10 after $50\%$ and $75\%$ of the training. 
In all the experiments, the weight decay is set to $1e^{-4}$, and a mini-batch size of 32 per agent is used.
We use an SGD optimizer with the Nesterov momentum (momentum coefficient = 0.9).
The stopping criteria is a fixed number of epochs. The experiments on CIFAR-10 are run for 200 epochs, CIFAR-100, Fashion MNIST and Imagenette for 100 epochs, and  ImageNet for 50 epochs. 
Note that DSGDm-N indicates Decentralized Stochastic Gradient Descent with Nesterov momentum. 
We report the test accuracy of the consensus model averaged over three randomly chosen seeds. 
A detailed description of the setup and hyperparameters for each experiment is presented in the Appendix.~\ref{apx:dl}.

\subsection{Decentralized Deep Learning Results}

We evaluate the efficiency of AdaG-SGD with the help of an exhaustive set of experiments. AdaG-SGD is compared with DeepSqueeze \cite{deepsqueeze} and CHOCO-SGD \cite{choco-sgd} to demonstrate that the proposed method outperforms the current state-of-the-art. Table.~\ref{tab:cf10} shows the average test accuracy for training the ResNet-20 model on the CIFAR-10 dataset with Top-K sparsification ($90\%, 99\%$ sparsification) and uniform quantization ($8,4,2$ bit quantization) over a ring topology of 16 and 32 agents. 
We observe that AdaG-SGD consistently outperforms CHOCO-SGD for all graph sizes and compressors. We notice an improvement of $0.41-1.78\%$ in the case of sparsification and $0-2.36\%$ for quantization.

\begin{table}[ht]
\caption{Test accuracy of different decentralized algorithms evaluated on CIFAR-10, with different communication compressors for various models over ring topologies. The results are averaged over three seeds where the standard deviation is indicated. $0\%$ sparsification indicates the full communication baseline.}
\label{tab:cf10}
\small
\begin{center}
\resizebox{1.0\columnwidth}{!}{
\begin{tabular*}{\textwidth}{cl @{\extracolsep{\fill}}*{3}{c}}
\hline
\multirow{ 2}{*}{Agents ($n$)} &\multirow{ 2}{*}{Method}& \multicolumn{3}{c}{Top-k Sparsification} \\
\cline{3-5}  
& & $0\%$ (full-comm.) & $90\%$ & $99\%$\\
 \hline
\multirow{5}{*}{$16$} & transmitted data/epoch (MB) & 205 & 30.7 & 3.09 \\
 & DeepSqueeze \cite{deepsqueeze} &  \multirow{3}{*}{$89.68 \pm 0.07$} &  $86.99 \pm 0.24$ & $84.19 \pm 0.49$ \\
 & CHOCO-SGD \cite{choco-sgd} &  & $89.04 \pm 0.24$ & $86.81 \pm 0.26$ \\
 & AdaG-SGD (ours)  & & $\mathbf{89.91} \pm 0.25$ & $\mathbf{87.44} \pm 0.31$  \\
 \hline
\multirow{5}{*}{$32$} & transmitted data/epoch (MB)  & 102 & 15.3 & 1.55 \\
 & DeepSqueeze \cite{deepsqueeze} &  \multirow{3}{*}{$88.00 \pm 0.13$} &  $81.64 \pm 0.53$ & $71.42 \pm 0.99$ \\
 & CHOCO-SGD \cite{choco-sgd} &  &$86.23 \pm 0.29$ & $81.76 \pm 0.71$ \\
 & AdaG-SGD (ours) &  & $\mathbf{88.01} \pm 0.26$ & $\mathbf{82.17} \pm 0.39$  \\
 \hline
 \hline
\multirow{ 2}{*}{Agents ($n$)} &\multirow{ 2}{*}{Method} & \multicolumn{3}{c}{Quantization} \\
 \cline{3-5} 
& & 8-bit & 4-bit & 2-bit \\
 \hline
\multirow{5}{*}{$16$} & transmitted data/epoch (MB)  & 51.2 & 25.6 &  12.8\\
 & DeepSqueeze \cite{deepsqueeze} &  $88.56  \pm 0.11$ &  $86.07 \pm 0.25$ & $84.17 \pm 0.24$ \\
 & CHOCO-SGD \cite{choco-sgd} & $\mathbf{90.06} \pm 0.06$ & $88.39 \pm 0.04$ & $86.29 \pm 0.59$ \\
 & AdaG-SGD (ours) & $89.83 \pm 0.26$ & $\mathbf{89.40} \pm 0.31$ & $\mathbf{86.48} \pm 0.17$  \\
 \hline
\multirow{5}{*}{$32$} & transmitted data/epoch (MB)  & 25.6 & 12.8 & 6.40 \\
 & DeepSqueeze \cite{deepsqueeze} &  $84.57 \pm 0.24$ &  $80.36 \pm 0.52$ & $68.46 \pm 0.18$ \\
 & CHOCO-SGD \cite{choco-sgd} & $88.36 \pm 0.14$ & $84.31 \pm 0.60$ & $77.81 \pm 0.31$ \\
 & AdaG-SGD (ours) & $\mathbf{88.77} \pm 0.03$ & $\mathbf{86.49} \pm 0.20$ & $\mathbf{80.17} \pm 0.19$  \\
 \hline
\end{tabular*}
}
\end{center}
\end{table}

\begin{table}[ht]
\caption{Average test accuracy of different decentralized algorithms evaluated on CIFAR-10 dataset trained on ResNet-20 over various graph topologies and top-k sparsification compressor. DSGDm-N indicates the full communication baseline.}
\label{tab:topologies}
\small
\begin{center}
\begin{tabular}{cccccc}
\hline
\multirow{ 2}{*}{Method}&\multicolumn{2}{c}{Dyck (32 agents)} & &\multicolumn{2}{c}{Torus (32 agents)}\\
\cline{2-3} \cline{5-6}
 &$90\%$   &$99\%$ &&$90\%$  &$99\%$\\
\hline
DSGDm-N \cite{d-psgd}  & \multicolumn{2}{c}{$88.39 \pm 0.50$}  & & \multicolumn{2}{c}{$88.48 \pm 0.29$}  \\
CHOCO-SGD \cite{choco-sgd} & $ 86.71 \pm  0.22$ & $82.41 \pm  0.81$ & & $87.46  \pm  0.28$ & $82.62 \pm 0.78$ \\
AdaG-SGD (ours) & $\mathbf{88.38} \pm 0.12$ & $\mathbf{83.21} \pm  0.40$ & & $\mathbf{88.36} \pm 0.50$ & $\mathbf{83.59} \pm 0.29$\\
\hline
\end{tabular}
\end{center}
\end{table}

We present the experimental results on various graph topologies and datasets to demonstrate the scalability and generalizability of AdaG-SGD.
We train the CIFAR-10 dataset on ResNet-20 architecture over the Dyck and Torus graphs to demonstrate the impact of connectivity on the proposed technique. As shown in Table.~\ref{tab:topologies}, we achieve  $0.8-1.7\%$ performance gains with varying spectral gap.
Further, we evaluate AdaG-SGD on various image datasets such as Fashion MNIST, and Imagenette and on challenging datasets such as CIFAR-100 and ImageNet. Table.~\ref{tab:datasets} shows that AdaG-SGD outperforms CHOCO-SGD by $0.13-1\%$ across various datasets. The results on ImageNet, large scale vision dataset, are shown in Table.~\ref{tab:imagenet} where AdaG-SGD improves the performance by $0.8-1\%$.
Therefore, we conclude that the proposed AdaG-SGD method reaps the benefits of adaptive consensus step-size in communication-efficient decentralized learning setups improving the performance by $\sim 0.1-2\%$. 

\begin{table}[ht]
\caption{Average test accuracy of different decentralized algorithms evaluated on various datasets using top-k sparsification compressor over ring topology of 16 agents. DSGDm-N indicates the full communication baseline.
}
\label{tab:datasets}
\small
\begin{center}
\resizebox{1.0\columnwidth}{!}{
\begin{tabular}{lcccccc}
\hline
 \multirow{2}{*}{Method} & \multicolumn{2}{c}{Fashion MNIST (LeNet-5)} & \multicolumn{2}{c}{CIFAR-100 (ResNet-20)} & \multicolumn{2}{c}{Imagenette (MobileNet-V2)}\\
 \cline{2-7}
  & $90\%$   & $99\%$ & $90\%$  & $99\%$ & $90\%$   & $99\%$ \\
\hline
 DSGDm-N \cite{d-psgd} &\multicolumn{2}{c}{$90.70 \pm 0.13$}  & \multicolumn{2}{c}{$59.96 \pm 1.14$}  & \multicolumn{2}{c}{$79.52 \pm 1.21$}  \\
 CHOCO-SGD \cite{choco-sgd} & $90.45 \pm 0.19$ & $89.87 \pm 0.04$ & $54.70 \pm 0.16$ & $46.99 \pm 0.40$ & $71.52 \pm 2.06$ & $66.65 \pm  1.80$\\
 AdaG-SGD (ours) & $\mathbf{90.58} \pm 0.29$ & $\mathbf{90.24} \pm 0.13 $ & $\mathbf{55.62} \pm 0.22$ & $\mathbf{47.72} \pm 0.14$ & $\mathbf{72.52} \pm 2.08$ & $\mathbf{67.38} \pm 0.34$\\
\hline
\end{tabular}
}
\end{center}
\end{table}

\begin{table}[ht]
\caption{Test accuracy of ImageNet trained on ResNet-18 architecture over a ring graph of 16 agents.
}
\label{tab:imagenet}
\small
\begin{center}
\begin{tabular}{cccc}
\hline
Graph & Method & $90\%$ & $99\%$\\
\hline
  \multirow{3}{*}{Ring} & DSGDm-N (full comm.) & \multicolumn{2}{c}{$65.62 \pm 0.03$} \\
& CHOCO-SGD \cite{choco-sgd} & $63.88 \pm 0.08$& $57.33 \pm 0.07$ \\
 &  AdaG-SGD (ours) & $\mathbf{64.68} \pm 0.10$ & $\mathbf{58.33} \pm 0.10$\\
\hline
\end{tabular}
\end{center}
\end{table}

\subsection{Ablation Study}

\begin{figure}[ht]
\centering     
\subfigure[$90\%$ sparsification, ResNet-20]
{\label{fig:agents}
\includegraphics[width=45mm]{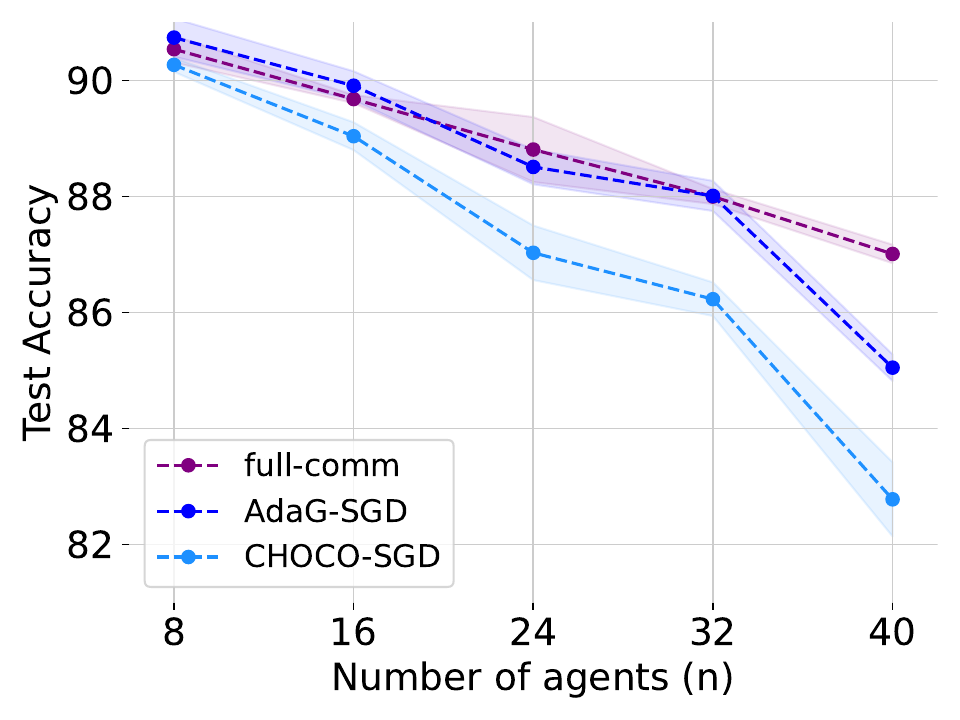}}
\subfigure[$90\%$ sparsification, $n=16$]
{\label{fig:depth}
\includegraphics[width=45mm]{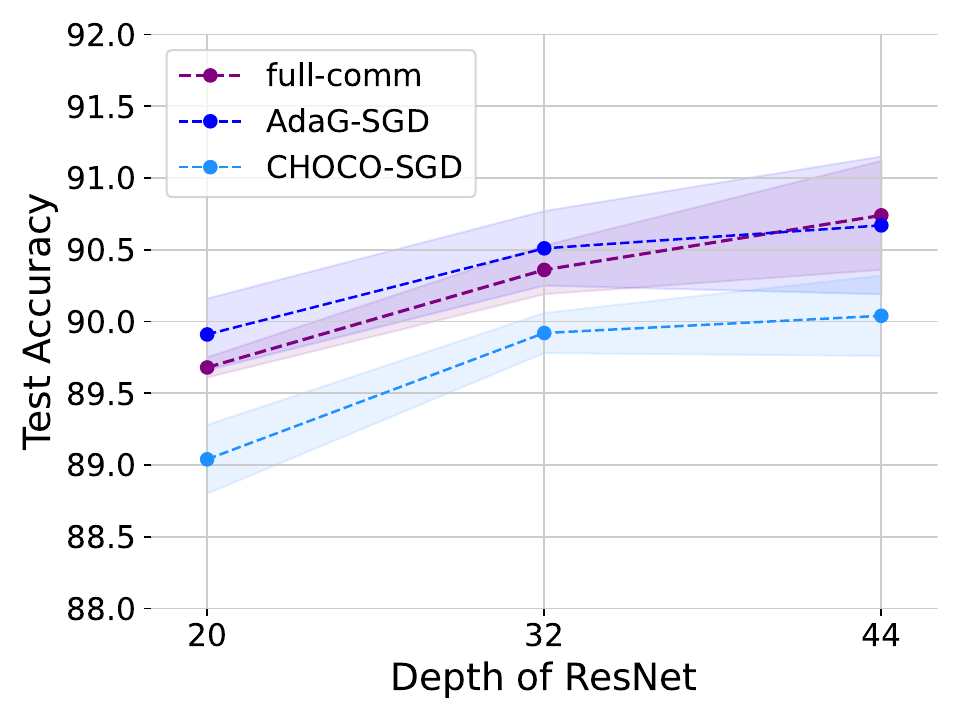}}
\subfigure[$n=16$, ResNet-20]
{\label{fig:beta}
\includegraphics[width=45mm]{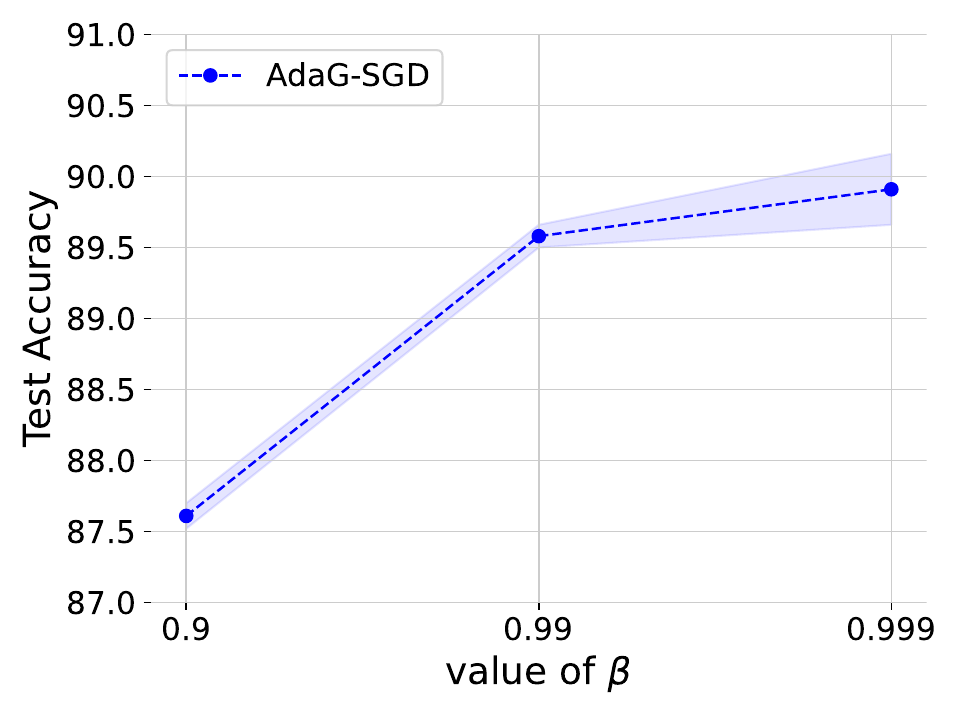}}
\caption{Ablation study on the hyper-parameter $\beta$, number of agents $n$ and model size. The test accuracy is reported for the CIFAR-10 dataset trained on ResNet architecture over ring topology.}
\label{fig:ablation}
\end{figure}

We evaluate the scalability of the proposed method in decentralized settings by training CIFAR-10 on varying the size of the ring topology between 8 and 40 as shown in Figure.~\ref{fig:agents}. The results show that AdaG-SGD consistently outperforms the CHOCO-SGD baseline across different graph sizes by an average improvement of $1\%$. Note that AdaG-SGD performs better than the full communication baseline for smaller graph structures. The regularization effect of the communication compression and adaptive consensus rate are the potential causes of this behavior. 
We then evaluate the proposed AdaG-SGD on the varying depth of ResNet architecture with ring topology of 16 agents as shown in Figure.~\ref{fig:depth}. We observe that the proposed method consistently outperforms the baseline over varying ResNet depth by an average improvement of $1.26\%$. The improvement in performance with AdaG-SGD is more prominent in graphs with high spectral gap i.e., larger graph topologies. 
Furthermore, Figure.~\ref{fig:beta} illustrates the effect of the moving average coefficient $\beta$ on the test accuracy of AdaG-SGD.

\section{Limitations}

This section discusses three potential limitations of the proposed AdaG-SGD method: 
1) AdaG-SGD assumes the mixing matrix to be doubly stochastic and symmetric. Therefore, the proposed method is not compatible with time-varying and directed graph topologies. The impact of adaptive consensus step-size on directed and time-varying graphs can be analyzed by combining AdaGossip with stochastic gradient push (SGP) \cite{sgp}. 
2) AdaG-SGD does not eliminate the tuning of hyper-parameter $\gamma$. Similar to other decentralized learning algorithms with communication compression, $\gamma$ needs to be tuned based on the compression ratio.
3) AdaG-SGD requires additional memory and computing to estimate the second raw moment of gossip-error. 
Further, we leave the theoretical formulation and analysis of the convergence rate of AdaGossip and AdaG-SGD as a future research direction.

\section{Conclusion}

Communication-efficient decentralized learning is the key to launching ML training on edge devices and thereby efficiently leveraging the humongous amounts of user-generated private data.
In this paper, we propose AdaGossip, a novel decentralized algorithm designed to improve performance with communication compression through adaptive consensus step-size. Further, we present AdaG-SGD, the adaptation of AdaGossip to decentralized machine learning.
The exhaustive set of experiments on different model architectures, datasets, and topologies demonstrates the superior performance of the proposed algorithm compared to the current state-of-the-art.

\section*{Acknowledgements}

This work was supported in part by, the Center for the Co-Design of Cognitive Systems (COCOSYS), a DARPA-sponsored JUMP center of Semiconductor Research Corporation (SRC).

{
\bibliographystyle{ieee_fullname}
\bibliography{egbib}

\begin{thebibliography}{10}\itemsep=-1pt

\bibitem{lp}
Sai~Aparna Aketi, Sangamesh Kodge, and Kaushik Roy.
\newblock Low precision decentralized distributed training over iid and non-iid data.
\newblock {\em Neural Networks}, 2022.

\bibitem{sparsepush}
Sai~Aparna Aketi, Amandeep Singh, and Jan Rabaey.
\newblock Sparse-push: Communication-\& energy-efficient decentralized distributed learning over directed \& time-varying graphs with non-iid datasets.
\newblock {\em arXiv preprint arXiv:2102.05715}, 2021.

\bibitem{sgp}
Mahmoud Assran, Nicolas Loizou, Nicolas Ballas, and Mike Rabbat.
\newblock Stochastic gradient push for distributed deep learning.
\newblock In {\em International Conference on Machine Learning}, pages 344--353. PMLR, 2019.

\bibitem{sgd}
L{\'e}on Bottou.
\newblock Large-scale machine learning with stochastic gradient descent.
\newblock In {\em Proceedings of COMPSTAT'2010}, pages 177--186. Springer, 2010.

\bibitem{vit}
Alexey Dosovitskiy, Lucas Beyer, Alexander Kolesnikov, Dirk Weissenborn, Xiaohua Zhai, Thomas Unterthiner, Mostafa Dehghani, Matthias Minderer, Georg Heigold, Sylvain Gelly, Jakob Uszkoreit, and Neil Houlsby.
\newblock An image is worth 16x16 words: Transformers for image recognition at scale.
\newblock 2021.

\bibitem{adagrad}
John Duchi, Elad Hazan, and Yoram Singer.
\newblock Adaptive subgradient methods for online learning and stochastic optimization.
\newblock {\em Journal of machine learning research}, 12(7), 2011.

\bibitem{resnet}
Kaiming He, Xiangyu Zhang, Shaoqing Ren, and Jian Sun.
\newblock Deep residual learning for image recognition.
\newblock In {\em Proceedings of the IEEE conference on computer vision and pattern recognition}, pages 770--778, 2016.

\bibitem{imagenette}
Hamel Husain.
\newblock Imagenette - a subset of 10 easily classified classes from the imagenet dataset.
\newblock {\em https://github.com/fastai/imagenette}, 2018.

\bibitem{choco-sgd}
Anastasia Koloskova, Tao Lin, Sebastian~U Stich, and Martin Jaggi.
\newblock Decentralized deep learning with arbitrary communication compression.
\newblock {\em arXiv preprint arXiv:1907.09356}, 2019.

\bibitem{choco1}
Anastasia Koloskova, Sebastian Stich, and Martin Jaggi.
\newblock Decentralized stochastic optimization and gossip algorithms with compressed communication.
\newblock In {\em Proceedings of the 36th International Conference on Machine Learning}, volume~97 of {\em Proceedings of Machine Learning Research}, pages 3478--3487. PMLR, 09--15 Jun 2019.

\bibitem{konevcny2016federated}
Jakub Kone{\v{c}}n{\`y}, H~Brendan McMahan, Daniel Ramage, and Peter Richt{\'a}rik.
\newblock Federated optimization: Distributed machine learning for on-device intelligence.
\newblock 2016.

\bibitem{cifar}
Alex Krizhevsky, Vinod Nair, and Geoffrey Hinton.
\newblock Cifar (canadian institute for advanced research).
\newblock {\em http://www.cs.toronto.edu/~kriz/cifar.html}, 2014.

\bibitem{lenet}
Yann LeCun, L{\'e}on Bottou, Yoshua Bengio, and Patrick Haffner.
\newblock Gradient-based learning applied to document recognition.
\newblock {\em Proceedings of the IEEE}, 86(11):2278--2324, 1998.

\bibitem{d-psgd}
Xiangru Lian, Ce Zhang, Huan Zhang, Cho-Jui Hsieh, Wei Zhang, and Ji Liu.
\newblock Can decentralized algorithms outperform centralized algorithms? a case study for decentralized parallel stochastic gradient descent.
\newblock {\em Advances in Neural Information Processing Systems}, 30, 2017.

\bibitem{qgm}
Tao Lin, Sai~Praneeth Karimireddy, Sebastian Stich, and Martin Jaggi.
\newblock Quasi-global momentum: Accelerating decentralized deep learning on heterogeneous data.
\newblock In {\em Proceedings of the 38th International Conference on Machine Learning}, volume 139 of {\em Proceedings of Machine Learning Research}, pages 6654--6665. PMLR, 18--24 Jul 2021.

\bibitem{evonorm}
Hanxiao Liu, Andy Brock, Karen Simonyan, and Quoc Le.
\newblock Evolving normalization-activation layers.
\newblock In H. Larochelle, M. Ranzato, R. Hadsell, M.~F. Balcan, and H. Lin, editors, {\em Advances in Neural Information Processing Systems}, volume~33, pages 13539--13550. Curran Associates, Inc., 2020.

\bibitem{moniqua}
Yucheng Lu and Christopher De~Sa.
\newblock Moniqua: Modulo quantized communication in decentralized {SGD}.
\newblock In {\em Proceedings of the 37th International Conference on Machine Learning}, volume 119 of {\em Proceedings of Machine Learning Research}, pages 6415--6425. PMLR, Jul 2020.

\bibitem{nedic2020distributed}
Angelia Nedic.
\newblock Distributed gradient methods for convex machine learning problems in networks: Distributed optimization.
\newblock {\em IEEE Signal Processing Magazine}, 37(3):92--101, 2020.

\bibitem{gpt4}
OpenAI.
\newblock Gpt-4 technical report.
\newblock {\em arXiv preprint arXiv:2303.08774}, 2023.

\bibitem{mobilnetv2}
Mark Sandler, Andrew Howard, Menglong Zhu, Andrey Zhmoginov, and Liang-Chieh Chen.
\newblock Mobilenetv2: Inverted residuals and linear bottlenecks.
\newblock In {\em Proceedings of the IEEE conference on computer vision and pattern recognition}, pages 4510--4520, 2018.

\bibitem{quant_sgp}
Hossein Taheri, Aryan Mokhtari, Hamed Hassani, and Ramtin Pedarsani.
\newblock Quantized decentralized stochastic learning over directed graphs.
\newblock In {\em Proceedings of the 37th International Conference on Machine Learning}, volume 119, pages 9324--9333, 2020.

\bibitem{deepsqueeze}
Hanlin Tang, Xiangru Lian, Shuang Qiu, Lei Yuan, Ce Zhang, Tong Zhang, and Ji Liu.
\newblock Deepsqueeze: Decentralization meets error-compensated compression.
\newblock {\em arXiv preprint arXiv:1907.07346}, 2019.

\bibitem{powersgd}
Thijs Vogels, Sai~Praneeth Karimireddy, and Martin Jaggi.
\newblock Powersgd: Practical low-rank gradient compression for distributed optimization.
\newblock {\em Advances in Neural Information Processing Systems}, 32, 2019.

\bibitem{powergossip}
Thijs Vogels, Sai~Praneeth Karimireddy, and Martin Jaggi.
\newblock Powergossip: Practical low-rank communication compression in decentralized deep learning.
\newblock {\em arXiv preprint arXiv:2008.01425}, 2020.

\bibitem{fmnist}
Han Xiao, Kashif Rasul, and Roland Vollgraf.
\newblock Fashion-mnist: a novel image dataset for benchmarking machine learning algorithms.
\newblock {\em arXiv preprint arXiv:1708.07747}, 2017.

\bibitem{gossip}
Lin Xiao and Stephen Boyd.
\newblock Fast linear iterations for distributed averaging.
\newblock {\em Systems \& Control Letters}, 53(1):65--78, 2004.

\bibitem{beer}
Haoyu Zhao, Boyue Li, Zhize Li, Peter Richt{\'a}rik, and Yuejie Chi.
\newblock Beer: Fast $ o (1/t) $ rate for decentralized nonconvex optimization with communication compression.
\newblock {\em Advances in Neural Information Processing Systems}, 35:31653--31667, 2022.

\end{thebibliography}
}
%

\clearpage
\appendix
\clearpage

\section{Decentralized Learning Setup}
\label{apx:dl}
All our experiments were conducted on a system with Nvidia GTX 1080ti card with 4 GPUs except for ImageNet simulations. We used NVIDIA A100 card with 4 GPUs for ImageNette and ImageNet simulations.

\subsection{Datasets}
\label{apx:datasets}
In this section, we give a brief description of the datasets used in our experiments. We use a diverse set of datasets each originating from a different distribution of images to show the generalizability of the proposed techniques.

\textbf{CIFAR-10:} 
CIFAR-10 \cite{cifar} is an image classification dataset with 10 classes. The image samples are colored (3 input channels) and have a resolution of $32 \times 32$. 
There are $50,000$ training samples with $5000$ samples per class and $10,000$ test samples with $1000$ samples per class.

\textbf{CIFAR-100:} 
CIFAR-100 \cite{cifar} is an image classification dataset with 100 classes. The image samples are colored (3 input channels) and have a resolution of $32 \times 32$. There are $50,000$ training samples with $500$ samples per class and $10,000$ test samples with $100$ samples per class. CIFAR-100 classification is a harder task compared to CIFAR-10 as it has 100 classes with very few samples per class to learn from.

\textbf{Fashion MNIST:}
Fashion MNIST \cite{fmnist} is an image classification dataset with 10 classes. The image samples are in greyscale (1 input channel) and have a resolution of $28 \times 28$. There are $60,000$ training samples with $6000$ samples per class and $10,000$ test samples with $1000$ samples per class.

\textbf{Imagenette:}
Imagenette \cite{imagenette} is a 10-class subset of the ImageNet dataset. The image samples are colored (3 input channels) and have a resolution of $224 \times 224$. There are $9469$ training samples with roughly $950$ samples per class and $3925$ test samples. 

\textbf{ImageNet:}
ImageNet dataset spans 1000 object classes and contains 1,281,167 training images, 50,000 validation images, and 100,000 test images. The image samples are colored (3 input channels) and have a resolution of $224 \times 224$.

\subsection{Network Architecture}
\label{apx:arch}
We replace ReLU+BatchNorm layers of all the model architectures with EvoNorm-S0 as it was shown to be better suited for decentralized learning over non-IID distributions.

\textbf{ResNet-20:} For ResNet-20 \cite{resnet}, we use the standard architecture with $0.27M$ trainable parameters except that BatchNorm+ReLU layers are replaced by EvoNorm-S0.

\textbf{ResNet-18:} For ResNet-18 \cite{resnet}, we use the standard architecture with $11M$ trainable parameters except that BatchNorm+ReLU layers are replaced by EvoNorm-S0.

\textbf{LeNet-5:} For LeNet-5 \cite{lenet}, we use the standard architecture with $61,706$ trainable parameters.

\textbf{MobileNet-V2:} We use the the standard MobileNet-V2 \cite{mobilnetv2} architecture used for CIFAR dataset with $2.3M$ parameters except that BatchNorm+ReLU layers are replaced by EvoNorm-S0.

\subsection{Consensus Rate}

The constant consensus rate ($\gamma$) for CHOCO-SGD and the hyper-parameter $\gamma$ used to compute the adaptive consensus rate ($\gamma_t$) for AdaG-SGD are tuned using the validation dataset via grid search. Note that the moving average coefficient $\beta$ is set to $0.999$ in all the experiments. The final values of $\gamma$ used in the experiments are presented here.

\begin{table}[ht]
\caption{The value of $\gamma$ used in different decentralized algorithms evaluated on CIFAR-10, with different communication compressors for various models over ring topology (refer to Table.~\ref{tab:cf10}).}
\label{tab:cf10_hp}
\small
\begin{center}
\resizebox{1.0\columnwidth}{!}{
\begin{tabular*}{\textwidth}{cl @{\extracolsep{\fill}}*{3}{c}}
\hline
\multirow{ 2}{*}{Agents ($n$)} &\multirow{ 2}{*}{Method}& \multicolumn{3}{c}{Top-k Sparsification} \\
\cline{3-5}  
& & $0\%$ (full-comm.) & $90\%$ & $99\%$\\
 \hline
 \multirow{4}{*}{$16$}  & DeepSqueeze \cite{deepsqueeze} &  \multirow{3}{*}{$1$} &  $0.05$ & $0.01$ \\
 & CHOCO-SGD \cite{choco-sgd} &  & $0.2$ & $0.0375$ \\
 & AdaG-SGD (ours)  & & $0.01$ & $0.001$  \\
 \hline
\multirow{4}{*}{$32$}  & DeepSqueeze \cite{deepsqueeze} &  \multirow{3}{*}{$1$} &  $0.1$ & $0.02$ \\
 & CHOCO-SGD \cite{choco-sgd} &  &$0.2$ & $0.05$ \\
 & AdaG-SGD (ours) &  & $0.004$ & $0.0008$  \\
 \hline
 \hline
\multirow{ 2}{*}{Agents ($n$)} &\multirow{ 2}{*}{Method} & \multicolumn{3}{c}{Quantization} \\
 \cline{3-5} 
& & 8-bit & 4-bit & 2-bit \\
 \hline
  \multirow{4}{*}{$16$}  & DeepSqueeze \cite{deepsqueeze} &  $0.1$ &  $0.02$ & $0.01$ \\
 & CHOCO-SGD \cite{choco-sgd} & $0.7$ & $0.1$ & $0.025$ \\
 & AdaG-SGD (ours) & $0.008$ & $0.002$ & $0.0008$  \\
 \hline
 \multirow{4}{*}{$32$} & DeepSqueeze \cite{deepsqueeze} &  $0.1$ &  $0.08$ & $0.03$ \\
 & CHOCO-SGD \cite{choco-sgd} & $0.8$ & $0.1$ & $0.025$ \\
 & AdaG-SGD (ours) & $0.02$ & $0.002$ & $0.0008$  \\
 \hline
\end{tabular*}
}
\end{center}
\end{table}

\begin{table}[ht]
\caption{The value of $\gamma$ used in different decentralized algorithms evaluated on CIFAR-10 dataset trained on ResNet-20 over various graph topologies and top-k sparsification compressor (refer to Table.~\ref{tab:topologies}).}
\label{tab:topologies_hp}
\small
\begin{center}
\begin{tabular}{cccccc}
\hline
\multirow{ 2}{*}{Method}&\multicolumn{2}{c}{Dyck (32 agents)} & &\multicolumn{2}{c}{Torus (32 agents)}\\
\cline{2-3} \cline{5-6}
 &$90\%$   &$99\%$ &&$90\%$  &$99\%$\\
\hline
DSGDm-N \cite{d-psgd}  & \multicolumn{2}{c}{$1$}  & & \multicolumn{2}{c}{$1$}  \\
CHOCO-SGD \cite{choco-sgd} & $ 0.15$ & $0.03$ & & $0.15$ & $0.03$ \\
AdaG-SGD (ours) & $0.004$ & $0.0008$ & & $0.004$ & $0.001$\\
\hline
\end{tabular}
\end{center}
\end{table}

\begin{table}[ht]
\caption{The value of $\gamma$ used in different decentralized algorithms evaluated on various datasets using top-k sparsification compressor over ring topology of 16 agents (refer to Table.~\ref{tab:datasets}). }
\label{tab:datasets_hp}
\small
\begin{center}
\resizebox{1.0\columnwidth}{!}{
\begin{tabular}{lcccccc}
\hline
 \multirow{2}{*}{Method} & \multicolumn{2}{c}{Fashion MNIST (LeNet-5)} & \multicolumn{2}{c}{CIFAR-100 (ResNet-20)} & \multicolumn{2}{c}{Imagenette (MobileNet-V2)}\\
 \cline{2-7}
  & $90\%$   & $99\%$ & $90\%$  & $99\%$ & $90\%$   & $99\%$ \\
\hline
 DSGDm-N \cite{d-psgd} &\multicolumn{2}{c}{$1$}  & \multicolumn{2}{c}{$1$}  & \multicolumn{2}{c}{$1$}  \\
 CHOCO-SGD \cite{choco-sgd} & $0.1$ & $0.01$ & $0.2$ & $0.04$ & $0.1$ & $0.06$\\
 AdaG-SGD (ours) & $0.002$ & $0.001$ & $0.01$ & $0.001$ & $0.005$ & $0.0003$\\
\hline
\end{tabular}
}
\end{center}
\end{table}

\begin{table}[ht]
\caption{The value of $\gamma$ used to train ImageNet on ResNet-18 architecture over a ring graph of 16 agents (refer to Table.~\ref{tab:imagenet}).}
\label{tab:imagenet_hp}
\small
\begin{center}
\begin{tabular}{cccc}
\hline
Graph & Method & $90\%$ & $99\%$\\
\hline
  \multirow{3}{*}{Ring} & DSGDm-N (full comm.) & \multicolumn{2}{c}{$1$} \\
& CHOCO-SGD \cite{choco-sgd} & $0.3$& $ 0.03 $ \\
 &  AdaG-SGD (ours) & $0.001$ & $  0.0001 $\\
\hline
\end{tabular}
\end{center}
\end{table}

\end{document}